# Using Chaos in Grey Wolf Optimizer and Application to Prime Factorization

Harshit Mehrotra and Dr. Saibal K. Pal

**Abstract** The Grey Wolf Optimizer (GWO) is a swarm intelligence meta-heuristic algorithm inspired by the hunting behaviour and social hierarchy of grey wolves in nature. This paper analyses the use of chaos theory in this algorithm to improve its ability to escape local optima by replacing the key parameters by chaotic variables. The optimal choice of chaotic maps is then used to apply the Chaotic Grey Wolf Optimizer (CGWO) to the problem of factoring a large semi prime into its prime factors. Assuming the number of digits of the factors to be equal, this is a computationally difficult task upon which the RSA-cryptosystem relies. This work proposes the use of a new objective function to solve the problem and uses the CGWO to optimize it and compute the factors. It is shown that this function performs better than its predecessor for large semi primes and CGWO is an efficient algorithm to optimize it.

## 1 Introduction

In the world of mathematical optimization and computational intelligence, solving NP-hard problems is a major challenge as no exact or complete methods exist which can solve these problems in polynomial time. These methods incur huge memory and runtime costs. A solution is to compromise on the chances of getting the correct solution of the problem by using population-based meta-heuristics. They are low

Harshit Mehrotra
Department of Computer Science and Engineering
Indian Institute of Technology (Banaras Hindu University) Varanasi, India
e-mail: harshit.mehrotra.cse15@iitbhu.ac.in

Dr. Saibal K. Pal
Scientific Analysis Group (SAG)
Defence Research and Development Organization (DRDO), New Delhi, India
e-mail: skptech@yahoo.com

on computation time and may also hold lower memory requirements. These meta-heuristics do not need gradient-related information. They begin with a set of candidate solutions which are then improved over the course of the runtime to achieve the true solution of the problem.

A branch of population based meta-heuristics that is very popular for mathematical optimization is swarm intelligence (SI). In the words of Bonabeau et al., it is The emergent collective intelligence of groups of simple agents. [2] These algorithms usually draw inspiration from the behaviour of organisms in nature. A number of such nature-inspired SI algorithms have been devised in order to solve problems of continuous optimization as well as combinatorial optimization (like traveling salesman problem, knapsack problem, vehicle routing problem). Some of these are the Particle Swarm Optimization (PSO) [6], Firefly Algorithm [20], Ant Colony Optimization (ACO) [4], Artificial Bee Colony (ABC) [10], Bat Algorithm [21] to name a few.

Another such algorithm that was found to be very efficient in solving continuous optimization problems is the Grey Wolf Optimizer (GWO) [13]. It is based on the social hierarchy and hunting behaviour of grey wolves. This meta-heuristic has some tendency to avoid local optimum and is also efficient in moving towards the true solution. [22] However, since it advances towards exploitation, it is not always good for global search. Here, we have tried to eradicate this shortcoming of the GWO by using chaotic variables. Chaotic dynamics find a major application in optimization meta-heuristics to solve the problem of local optimum convergence [19]. This has been successfully proven by application to many SI algorithms like Firefly algorithm [8], Bat algorithm [9] and PSO [1]. We apply chaos on the key parameters of GWO and compare the performance of the Chaotic GWO (CGWO) with the standard GWO using 6 benchmark functions.

We have then explored the efficiency of the CGWO to solve discrete optimization problems by applying it to the problem of factorizing a product of two large prime numbers. Such numbers, known as semi primes are very difficult to factorize when the factors are of almost equal number of digits. [3] No exact methods exist to solve this problem in polynomial time. Hence, some SI-based techniques have been used to solve this problem using certain objective functions [14, 3, 11]. Here, we propose the use of another objective function and compare it's performance with a previously used function along with optimization using CGWO.

The rest of the paper goes into the details of each point discussed above. Section 2 discusses the standard GWO meta-heuristic, followed by a detailed discussion of our model of the CGWO in Section 3. Various aspects of the prime factorization problem have been discussed in Section 4. Results of testing on the benchmark functions and semi primes are given in Section 5. Section 6 gives concluding remarks of this study.

## 2 The Grey Wolf Optimizer

Grey wolves are among the predators that form the top of the food chain. They have a strong social structure which is followed while hunting. In decreasing order of dominance, wolves in a pack are classified as alpha ($\alpha$), beta ($\beta$), delta ($\delta$) and omega ($\omega$) wolves. The hunting process is also divided into three phases [17]: tracking, chasing the prey ; encircling it and attacking it. Inspired by these properties of grey wolves, the Grey Wolf Optimizer was given by S. Mirjalilli et al. in 2014 [13]. In the mathematical model of the GWO, the fittest solution is labeled as the alpha ($\alpha$), followed by the beta ($\beta$) and the delta ($\delta$) which are the second and third fittest solutions, respectively. All other solutions are omegas ($\omega$) and follow the other three kinds. The process of encircling the prey is modeled by calculating a distance vector and using it to update the position of a wolf. The hunt is usually guided by the alpha and occasionally by the beta and delta. Eliminating this uncertainty for the purpose of mathematical modeling, it is assumed the best 3 solutions have better knowledge about the optimum and all other solutions are updated according to the positions of the $\alpha$, $\beta$ and $\delta$ [13]. All the above discussed operations are formulated for an agent with position vector **X** as follows :

$$\mathbf{D}_\alpha = |\mathbf{C}_1 \cdot \mathbf{X}_\alpha - \mathbf{X}|, \ \mathbf{D}_\beta = |\mathbf{C}_2 \cdot \mathbf{X}_\beta - \mathbf{X}|, \ \mathbf{D}_\delta = |\mathbf{C}_3 \cdot \mathbf{X}_\delta - \mathbf{X}| \quad (1)$$

$$\mathbf{X}_1 = \mathbf{X}_\alpha - \mathbf{A}_1 \cdot \mathbf{D}_\alpha, \ \mathbf{X}_2 = \mathbf{X}_\beta - \mathbf{A}_2 \cdot \mathbf{D}_\beta, \ \mathbf{X}_3 = \mathbf{X}_\delta - \mathbf{A}_3 \cdot \mathbf{D}_\delta \quad (2)$$

$$\mathbf{X}(t+1) = \frac{\mathbf{X}_1 + \mathbf{X}_2 + \mathbf{X}_3}{3} \quad (3)$$

The vectors **A** and **C** are defined as:

$$\mathbf{A} = 2\mathbf{a} \cdot \mathbf{r}_1 - \mathbf{a} \quad (4)$$

$$\mathbf{C} = 2\mathbf{r}_2 \quad (5)$$

The components of **a** are uniformly decreased from 2 to 0 over iterations. $\mathbf{r}_1$ and $\mathbf{r}_2$ are random vectors in [0,1]. A detailed discussion of the nature of search (explorative and exploitative) is presented in the next section where chaotic improvements to the GWO are suggested.

After having been proved effective for continuous optimization and some engineering problems [13], various other versions of the GWO have been proposed to solve various problems like optimizing the control parameters of a DC motor [12], feature extraction [7], training q-gaussian radial basis functional link nets [16], to name a few.

## 3 Chaos in Grey Wolf Optimizer

Chaos is a characteristic of any non-linear system. It is basically a bounded unstable behaviour that occurs in a deterministic non-linear system. Any chaotic system possesses the property of sensitive dependence on initial conditions, implying that the slightest change in the parameters or initial conditions can lead to a vast difference in the future behavious of the system [5].

### 3.1 Why Chaos?

As the optimization problem gets tougher with a large number of local optima (like multi-modal functions), the chances that a population-based meta-heuristic will get trapped in one such local optimum increases. Chaos has been used in recent times to solve this problem by developing chaotic optimization algorithms [19]. When used suitably, the pseudo-randomness, ergodicity and irregularity of chaotic variables helps algorithms alternate between exploration and exploitation, and hence avoid getting trapped in a local solution. Moreover, chaos is non-repetitive, thus enabling these methods to carry out overall searches at higher speeds than stochastic searches that depend on probabilities [5].

### 3.2 Chaotic Maps

In order to incorporate chaos in an optimization algorithm, we use one dimensional functions called chaotic maps which exhibit the property of 'sensitive dependence on initial conditions' and are used in place of key parameters of the algorithm. Some popular chaotic maps which have been used in this paper are [9]:

1. Gauss map :

$$x_{k+1} = \begin{cases} 0 & x_k = 0 \\ \frac{1}{x_k} \mod 1 & otherwise \end{cases} \qquad (6)$$

   It generates chaotic sequences in (0,1).
2. Logistic map :

$$x_{k+1} = ax_k(1-x_k) \qquad (7)$$

   It generates chaotic sequences in (0,1) provided that $x_0 \in (0,1)$ and that $x_0 \notin 0.0, 0.25, 0.75, 0.5, 1.0$. Here, we have used $a = 4$.
3. Chebyshev map :

$$x_{k+1} = cos(kcos^{-1}x_k) \qquad (8)$$

This map generates a chaotic sequence in (-1,1).

4. Iterative map :

$$x_{k+1} = sin(\frac{a\pi}{x_k})  \quad (9)$$

Here, $a \in (0,1)$ is a suitable parameter. The chaotic sequence lies in (-1,1).

5. Singer map:

$$x_{k+1} = \mu(7.86x_k - 23.31x_k^2 + 28.75x_k^3 - 13.3x_k^4) \quad (10)$$

Here, $\mu$ lies between 0.9 and 1.08.

6. Tent map:

$$x_{k+1} = \begin{cases} \frac{x_k}{0.7} & x_k < 0.7 \\ \frac{10}{3}(1-x_k) & x_k \geq 0.7 \end{cases} \quad (11)$$

It generates a chaotic sequence in (0,1).

7. Sinusoidal map:

$$x_{k+1} = ax_k^2 sin(\pi x_k) \quad (12)$$

This map also generates chaotic sequences in (0,1). When $a = 2.3$ and $x_0 = 0.7$, it simplifies as:

$$x_{k+1} = sin(\pi x_k) \quad (13)$$

## 3.3 Adding Chaos to the GWO

A lot of study has been done in the field of chaotic optimization algorithms by developing chaotic versions of algorithms by replacing control parameters or random variables by chaotic variables [1, 8, 9]. The main motive is to alternate between exploration and exploitation so that the agents don't get caught in a local optimum.

The key parameters in GWO are **A** and **C**. Since $\mathbf{r}_1$ has its elements in [0,1], the value of **A** can vary in $[-a,a]$. We have $\mathbf{C} = 2\mathbf{r}_2$ where $\mathbf{r}_2$ has its elements in [0,1], thus implying that **C** has its elements in [0,2].

*Introducing chaos in a*

From the position update equation, we get that the new position of a wolf is between the current position and the prey i.e. attacking it (exploitation) when $|A| < 1$ [13]. As $a$ is uniformly decreased from 2 to 0, once $a$ comes below 1 (after half the number of maximum iterations), the search will be invariably exploitative. In the first half, it changes between explorative and exploitative. To keep such a behaviour intact, we replace $a$ by a chaotic sequence normalized to give values in [1,2]. This makes sure that the search can have both natures throughout the iterations. fig. 1 and fig. 2 show the variation of *A* for a standard GWO and a chaotic GWO (CGWO) with Tent map used for *a*, respectively when run for 100 iterations. The red line indicates $a$, $-a$ and the blue line indicates $A$. The bold black line indicates the values 1 and -1. It can be seen clearly that the CGWO provides exploration at regular intervals throughout the search, thereby creating better balance. The standard GWO explores the search space for intervals only in the initial half.

*Introducing chaos in C*

The parameter $C$ assigns weight to the role of the prey in movement of the wolves [13]. When $C > 1$, the role of the prey is emphasized and is given less importance for $C < 1$. So, we redefine $C$ chaotically by replacing the random variable $r_2$ with a chaotic variable normalized between 0 and 1. Owing to the ergodicity and mixing property of chaos, replacing a random variable by a chaotic one is expected to give better results. Results of comparison of GWO and CGWO on some objective functions are given in Section 5.

## 4 The Prime Factorization Problem

A semi-prime number is a composite number obtained by multiplying two prime numbers. Factorizing such numbers is a difficult task, particularly when the the two factors are almost similar sized [3]. This can be understood by knowing that prime factorization is a one-way trapdoor function, which means that given two prime numbers $p$ and $q$, we can easily compute the corresponding semi-prime number $N$. However, it is very difficult to obtain $p$ and $q$ if we are given $N$. The difficulty of this problem is used in the RSA cryptosystem. The problem is NP-complete, so no polynomial time algorithms exists for it for non-quantum computers. The most efficient exact method developed is the General Number Field Sieve method [18] whose asymptotic time complexity is:

$$O(exp((\frac{64}{9})^{\frac{1}{3}}(\log b)^{\frac{2}{3}})) \qquad (14)$$

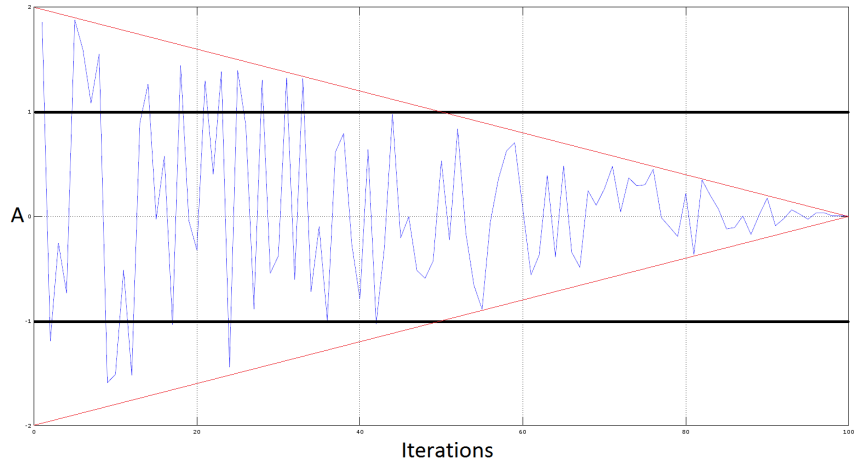

**Fig. 1** Variation of *A* in standard GWO

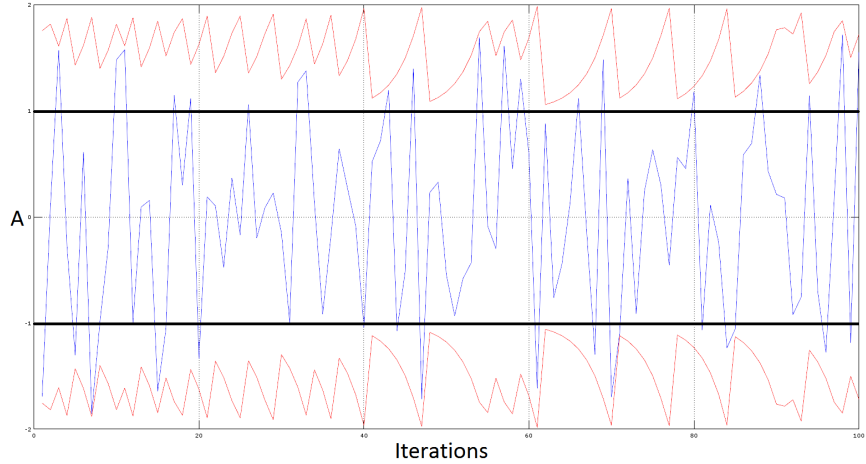

**Fig. 2** Variation of *A* with Tent map applied to *a*

Efforts have been made to explore the possibility of solving this problem using meta-heuristics [14, 3, 11]. Various techniques have been applied. Some of them have given promising results as well. We assume that both factors have equal number of digits and propose an improvement to the current methods.

## 4.1 Choosing an Objective Function

Solving any problem using meta-heuristics involves the optimization of an objective function. Prime factorization follows a similar procedure. Since the problem is very difficult, changes in the objective function can gave largely varying results. In fact, the main challenge of solving this problem is choosing a suitable objective function [15]. Initially the following two-dimensional function was used for the purpose [11]:

$$\textbf{minimize } f(x) = (x^2 - y^2) \mod N \qquad x, y \in [2, N-1] \qquad (15)$$

$$\textbf{constraint } (x \pm y) \mod N \neq 0 \qquad (16)$$

The factors are then calculated as the GCD of $x+y$ and $x-y$ with $N$. This function is highly chaotic with many local minima even for a small $N$. Moreover, the search space is very large. Thus a one dimensional function was introduced [3] :

$$\textbf{minimize } f(x) = N \mod x \qquad x \in (2, \sqrt{N}) \tag{17}$$

This function attains 0 when $x$ is a factor of $N$. It is the smaller factor as the upper bound is $\sqrt{N}$. The other factor is obtained by dividing $N$ by the obtained $x$. The lower bound becomes $10^{d-1}$ under the assumption that both factors have same number of digits ($d$ is number of digits in $\sqrt{N}$. The probability that a randomly generated solution is correct in eq. (15) is $\frac{1}{(N-1)^2}$ and in eq. (17) is more than $\frac{1}{(N-1)}$. eq. (17) also has several local minima, bu being 1-dimensional, its performance is better as seen in [3]. This motivates us to explore possibilities of a better objective function to solve the problem. One aim should be to narrow down the search space as well.

*Objective function used*

Consider a semi-prime $N$ with prime factors $p(<\sqrt{N})$ and $q(>\sqrt{N})$. Both will be greater than 2, hence their sum is an even integer.

$$p+q = 2n \tag{18}$$

Using $N = pq$ to eliminate $q$, we get:

$$p^2 - 2pn + N = 0 \tag{19}$$

Solving eq. (19), we get 2 solutions which are the values of $p$ and $q$.

$$p = n - \sqrt{n^2 - N} \text{ and } q = n + \sqrt{n^2 - N} \tag{20}$$

Since both the factors are integers, we need to find a positive integer $n$ such that the radical $\sqrt{n^2 - N}$ is also an integer. Next, we need to fix the bounds within which such an $n$ has to be searched.

We continue with our assumption that both factors have equal number of digits, $d$. This gives us:

$$10^{d-1} < p < \sqrt{N} \text{ and } \sqrt{N} < q < 10^d - 1 \tag{21}$$

Writing $p$ and $q$ as their difference from $\sqrt{N}$:

$$p = \sqrt{N} - x \qquad 0 < x < \sqrt{N} - 10^{d-1} \tag{22}$$

$$q = \sqrt{N} + y \qquad 0 < y < 10^d - 1 - \sqrt{N} \tag{23}$$

Using the above two equations and $N = pq$, we obtain the following relation:

$$xy = 2n\sqrt{N} - 2N \tag{24}$$

The mean $n$ can now be written as:

$$n = \sqrt{N} + \frac{y-x}{2} \quad (25)$$

We now use eq. (24) to get a quadratic equation in $x$ by substituting for $y$ in eq. (25). This gives us the following value of $n$:

$$n = \frac{x^2 - 2x\sqrt{N} + 2N}{2(\sqrt{N} - x)} \quad (26)$$

Using differentiation techniques, it can be calculated that the above expression will be increasing for the domain of $x$ given in eq. (22). Hence, the domain of $n$ comes out to be:

$$n \in (\sqrt{N}, \frac{10^{d-1} + \frac{N}{10^{d-1}}}{2}) \quad (27)$$

Using the domains of $x$, $y$ in eqs. (22) and (23) and the fact that $n > \sqrt{N}$ (AM-GM inequality), we obtain another range for $n$:

$$n \in (\sqrt{N}, \frac{\sqrt{N} + 10^d - 1}{2}) \quad (28)$$

The upper bound for searching $n$ will be the least of the ones obtained in eqs. (27) and (28) depending upon the $N$ used. So, the final objective function becomes:

$$\textbf{minimize } \{\sqrt{x^2 - N}\} \quad x \in (\sqrt{N}, \min(\frac{10^{d-1} + \frac{N}{10^{d-1}}}{2}, \frac{\sqrt{N} + 10^d - 1}{2})) \quad (29)$$

where $\{.\}$ refers to the fractional part function.

Results of optimizing this function using CGWO have been given and compared with those using with the function in eq. (17) in the next section.

## 5 Experiments and Results

For the first part of the experiments, the CGWO with chaos in $a$ (section 3.3) and $C$ (section 3.3 is compared with the standard GWO for performance on 5 minimization continuous benchmark functions. These are the Rastringin function ($f_1$), Ackley function ($f_2$), Sphere function ($f_3$), Goldstein-Price function ($f_4$) and Griewank function ($f_5$). 4 of these being multi-modal, test the explorative ability and the Sphere function tests the efficiency of exploitation. Parameters for these functions are given in table 1 and their definitions are given below. All tests are carried out in GNU Octave 4.0.2 on a 8GB, 2.20 GHz CPU laptop computer running Windows 8.

$$f_1(\mathbf{x}) = f(x_1, ..., x_n) = 10n + \sum_{i=1}^{n}(x_i^2 - 10\cos(2\pi x_i)) \quad (30)$$

$$f_2(\mathbf{x}) = -20.exp(-0.2\sqrt{\frac{1}{n}\sum_{i=1}^{n}x_i^2}) - exp(\frac{1}{d}\sum_{i=1}^{n}cos(2\pi x_i)) + 20 + e \quad (31)$$

$$f_3(\mathbf{x}) = \sum_{i=1}^{n}x_i^2 \quad (32)$$

$$f_4(x,y) = [1+(x+y+1)^2(19-14x+3x^2-14y+6xy+3y^2)][30+(2x-3y)^2(18-32x+12x^2+4y-36xy+27y^2)] \quad (33)$$

$$f_5(\mathbf{x}) = 1 + \sum_{i=1}^{n}\frac{x_i^2}{4000} - \prod_{i=1}^{n}cos(\frac{x_i}{\sqrt{i}}) \quad (34)$$

The number of search agents and maximum iterations are fixed at 30 and 500, respectively. Mean value and standard deviation are calculated for 30 runs of each type of chaotic map. The initial value of the chaotic variable is taken to be a random number between 0 and 1 to get an unbiased idea of the efficiency of the map. The chaotic maps used are the ones described in section 3.2. The GWO was proven to be better (or competitive, at least) for continuous optimization when compared to other popular meta-heuristics in [13]. Here, we have compared the CGWO with GWO and have evaluated the results using the mean, standard deviation (SD) and success ratio (SR). The success ratio is defined as the percentage of runs for which the solution is found successfully i.e. [9]

$$\sum_{d=1}^{D}(X_i^{obt} - X_i^*)^2 \leq (UB - LB) \times 10^{-4} \quad (35)$$

### 5.1 Results with chaos applied in $a$

As described in section 3.3, $a$ is replaced by a chaotic variable normalized in [0,1] and the results are given in tables 2 and 3. Moreover to give final exploitation, $a$ is given a value of 0.2 for the final 50 iterations, so as to give $A$ small values. It can be seen that in functions where the SR was already 100 ($f_2, f_3$), chaos helped in improving the mean value of the solution found. This improvement is of several orders in the case of $f_3$, a purely exploitation-intensive function. Performance for $f_4$ is the same, which was already perfect. Whereas, in $f_1$ and $f_5$, where the standard GWO failed to give correct answers every time, chaos improved the SR significantly with some maps and gave mean values comparable or better in those cases. Looking at both the tables, we can easily say that CGWO with the sinusoidal map is the best chaotic improvement in $a$.

**Table 1** Parameters for the benchmark functions used

| Definition | Dim | Domain | $f_{min}$ |
|---|---|---|---|
| $f_1(\mathbf{x})$ | 30 | [-5.12,5.12] | 0 |
| $f_2(\mathbf{x})$ | 30 | [-32,32] | 0 |
| $f_3(\mathbf{x})$ | 30 | [-100,100] | 0 |
| $f_4(x,y)$ | 2 | [-2,2] | 3 |
| $f_5(\mathbf{x})$ | 30 | [-600,600] | 0 |

## 5.2 Results with chaos applied in $C$

Next chaos is applied to $C$ as described in section 3.3. The results are tabulated in tables 4 and 5. The performance in case of $f_3$ is better in a few maps, but only by few orders. This is in contrast to the CGWO with chaos in $a$. This shows that a major part of exploitation is played by $a$, and not $C$. For $f_2$, most maps give improved means as compared to the standard GWO. Once again, the results are same for $f_4$. A number of maps give better success ratios and mean values of solutions for $f_1$ and $f_5$. From both the tables, it is inferred that CGWO with chaos in $C$ gives best performance with sinusoidal and iterative maps (almost the same between them).

**Table 2** Evaluation results for chaos in $a$

| Map | $f_1$ | | $f_2$ | | $f_3$ | | $f_4$ | | $f_5$ | |
|---|---|---|---|---|---|---|---|---|---|---|
| | Mean | SD | Mean | SD | Mean | SD | Mean | SD | Mean | SD |
| Gauss | 55.041 | 33.024 | 2.18E-14 | 3.01E-15 | 3.23E-50 | 8.74E-50 | 3 | 4.77E-05 | 0.0076 | 0.0084 |
| Logistic | 51.037 | 29.315 | 2.18E-14 | 2.72E-15 | 1.56E-51 | 4.31E-51 | 3 | 3.81E-05 | 0.0081 | 0.0081 |
| Sinusoidal | 11.0952 | 14.9742 | 1.47E-14 | 2.82E-15 | 5.72E-56 | 1.41E-55 | 3 | 2.54E-05 | 0.0026 | 0.0062 |
| Tent | 42.413 | 36.275 | 2.18E-14 | 2.59E-15 | 9.82E-51 | 4.54E-50 | 3 | 2.63E-05 | 0.0041 | 0.0044 |
| Singer | 24.487 | 29.45 | 1.47E-14 | 3.02E-15 | 7.74E-52 | 1.71E-51 | 3 | 3.64E-05 | 0.0027 | 0.0071 |
| Chebychev | 47.73 | 36.838 | 2.18E-14 | 3.53E-15 | 7.93E-52 | 1.56E-51 | 3 | 2.47E-05 | 0.0051 | 0.0084 |
| Iterative | 47.377 | 31.994 | 2.18E-14 | 3.15E-15 | 1.62E-47 | 5.16E-47 | 3 | 2.51E-05 | 0.0047 | 0.0076 |
| Standard | 13.955 | 10.079 | 8.57E-14 | 7.31E-15 | 3.94E-31 | 6.63E-31 | 3 | 2.90E-05 | 0.0046 | 0.0088 |

## 5.3 A different maximization problem

In order to test the efficiency of the CGWO, another function, quite a different one was also used. This function is quite smooth but there are plenty of local optima that yield values very close to the actual solution.

**Table 3** Success ratios with chaos in *a*

| Map | $f_1$ | $f_2$ | $f_3$ | $f_4$ | $f_5$ | Total SR |
|---|---|---|---|---|---|---|
| Gauss | 0 | 100 | 100 | 100 | 50 | 350 |
| Logistic | 0 | 100 | 100 | 100 | 50 | 350 |
| Sinusoidal | 52 | 100 | 100 | 100 | 84 | 436 |
| Tent | 0 | 100 | 100 | 100 | 90 | 390 |
| Singer | 40 | 100 | 100 | 100 | 87 | 427 |
| Chebychev | 0 | 100 | 100 | 100 | 70 | 370 |
| Iterative | 0 | 100 | 100 | 100 | 70 | 370 |
| Standard | 20 | 100 | 100 | 100 | 75 | 395 |

**Table 4** Evaluation results for chaos in *C*

| Map | $f_1$ | | $f_2$ | | $f_3$ | | $f_4$ | | $f_5$ | |
|---|---|---|---|---|---|---|---|---|---|---|
| | Mean | SD | Mean | SD | Mean | SD | Mean | SD | Mean | SD |
| Gauss | 31.541 | 16.724 | 7.46E-14 | 1.59E-14 | 1.78E-28 | 5.57E-28 | 3 | 1.42E-05 | 0.0085 | 0.0156 |
| Logistic | 12.138 | 10.181 | 3.64E-14 | 6.10E-15 | 7.89E-33 | 2.75E-32 | 3 | 1.22E-05 | 0.0022 | 0.0071 |
| Sinusoidal | 10.24 | 11.67 | 7.20E-14 | 9.12E-15 | 1.01E-29 | 1.039E-29 | 3 | 2.17E-06 | 0.0025 | 0.0068 |
| Tent | 35.124 | 15.031 | 1.02E-13 | 3.46E-14 | 6.54E-27 | 1.90E-26 | 3 | 1.05E-05 | 0.0067 | 0.0106 |
| Singer | 17.293 | 9.3137 | 1.59E-12 | 1.80E-12 | 1.05E-22 | 1.68E-22 | 3 | 2.72E-05 | 0.005 | 0.0083 |
| Chebychev | 20.801 | 21.197 | 3.50E-14 | 6.66E-15 | 6.67E-33 | 1.92E-32 | 3 | 1.12E-05 | 0.0037 | 0.008 |
| Iterative | 14.48 | 23.236 | 3.38E-14 | 5.08E-15 | 9.60E-34 | 2.30E-33 | 3 | 1.08E-05 | 0.0028 | 0.0076 |
| Standard | 13.955 | 10.079 | 8.57E-14 | 7.31E-15 | 3.94E-31 | 6.63E-31 | 3 | 2.90E-05 | 0.0046 | 0.00878 |

**Table 5** Success ratios with chaos in *C*

| Map | $f_1$ | $f_2$ | $f_3$ | $f_4$ | $f_5$ | Total SR |
|---|---|---|---|---|---|---|
| Gauss | 4 | 100 | 100 | 100 | 64 | 368 |
| Logistic | 17 | 100 | 100 | 100 | 90 | 407 |
| Sinusoidal | 33 | 100 | 100 | 100 | 87 | 420 |
| Tent | 0 | 100 | 100 | 100 | 67 | 367 |
| Singer | 10 | 100 | 100 | 100 | 67 | 377 |
| Chebychev | 20 | 100 | 100 | 100 | 80 | 400 |
| Iterative | 30 | 100 | 100 | 100 | 90 | 420 |
| Standard | 20 | 100 | 100 | 100 | 75 | 395 |

**Table 6** Results for the maximization problem

| Maps | Chaos in $a$ | | | Chaos in $C$ | | |
|---|---|---|---|---|---|---|
| | Mean | SD | SR | Mean | SD | SR |
| Gauss | 58941 | 291.79 | 0 | 59999.05 | 0.7377 | 60 |
| Logistic | 58927.5 | 276.327 | 0 | 59998.829 | 0.8612 | 53 |
| Sinusoidal | 58854.6 | 327.887 | 0 | 59999.322 | 0.6797 | 60 |
| Tent | 59999.386 | 0.75433 | 84 | 59999.007 | 1.6546 | 77 |
| Singer | 58708.8 | 369.906 | 0 | 59998.476 | 1.7212 | 50 |
| Chebychev | 59059 | 15.55 | 0 | 59999.031 | 0.8901 | 67 |
| Iterative | 59059 | 15.55 | 0 | 59998.633 | 1.5617 | 50 |
| Standard | 59998.128 | 1.6351 | 37 | 59998.128 | 1.6351 | 37 |

$$f(x) = f(s_1, x_2, x_3, x_4, x_5) = \prod_{i=1}^{5} x_i \mod 60000 \qquad x_i \in (1, 10) \qquad (36)$$

The problem is a maximization one with the true maximum value tending towards 60000. Keeping the parameters same, the results obtained are tabulated in table 6. It is observed that only the Tent map gives very good results for chaotic $a$ whereas all maps give average to good results for a chaotic $C$. Still, chaos has evidently shown great improvement to the standard performance.

## 5.4 Results on the prime factorization problem

The one dimensional objective function suggested in eq. (29) (F2) is now tested for its performance with the most successful function used till now (eq. (17)) i.e. F1. We carry on with our assumption that both factors have equal number of digits. The challenge for optimizing using CGWO is choosing a proper combination of maps. From the experiments done above, it can be inferred that sinusoidal, tent map are efficient for $a$ and sinusoidal, iterative map are efficient for $C$. For the functions now in concern, it was found from performing a few runs that a combination of sinusoidal ($a$), iterative ($C$) map works well for eq. (17) and that of sinusoidal map for both $a$, $C$ works well for eq. (29). We went ahead with these choices and the optimization results are presented in table 7. SR is the number of times, out of 30, the correct factors were computed within the maximum iterations. MI and SD are measured for the successful runs only.

In this problem of prime factorization, accuracy plays a bigger role than the time taken, provided that the difference in the latter is no very large. From the results obtained, it can be observed that our suggested function outperforms the currently used modular function in comparable number of iterations. The difference can be seen in a more clear way for the larger numbers. Following these results, this function can perhaps be used for further research even when number of digits of the factors a re not same by performing calculations to restrict the search space.

**Table 7** Comparison of objective functions for prime factorization

| Max Iter | Bits | N | Agents | F1 MI | F1 SD | F1 SR | F2 MI | F2 SD | F2 SR |
|---|---|---|---|---|---|---|---|---|---|
| 100 | 15 | 50759 | 30 | 15.57 | 13.772 | 30 | 16.96 | 17.909 | 30 |
| | | | 40 | 10.35 | 8.9017 | 30 | 12.44 | 11.012 | 30 |
| | | | 50 | 8.99 | 6.9143 | 30 | 9.5 | 9.2731 | 30 |
| 200 | 19 | 370627 | 30 | 53.08 | 59.67 | 30 | 46.83 | 42.6 | 30 |
| | | | 40 | 37.91 | 47.675 | 30 | 35.03 | 35.295 | 30 |
| | | | 50 | 29.35 | 32.444 | 30 | 24.89 | 24.006 | 30 |
| 500 | 24 | 10909343 | 80 | 74.45 | 67.096 | 30 | 106.44 | 105.031 | 30 |
| | | | 120 | 51.72 | 49.33 | 30 | 66.47 | 72.03 | 30 |
| | | | 160 | 35.67 | 35.261 | 30 | 50.8 | 49.549 | 30 |
| 500 | 25 | 29835457 | 80 | 133.93 | 114.68 | 29 | 112.467 | 88.834 | 30 |
| | | | 120 | 88.6 | 74.573 | 30 | 80.667 | 104.975 | 30 |
| | | | 160 | 67.8 | 49.353 | 30 | 77.167 | 71.852 | 30 |
| 1000 | 29 | 392913607 | 100 | 238.32 | 204.46 | 28 | 319.03 | 253.86 | 29 |
| | | | 300 | 138.1 | 102.036 | 30 | 171.333 | 144.276 | 30 |
| | | | 500 | 69.567 | 77.716 | 30 | 121.933 | 157.9 | 30 |
| 2000 | 33 | 5325280633 | 100 | 824.48 | 539.73 | 25 | 801.57 | 480.27 | 21 |
| | | | 300 | 485.9 | 489 | 29 | 465 | 463.13 | 29 |
| | | | 500 | 305.2 | 258.093 | 30 | 264.133 | 321.435 | 30 |
| 3000 | 35 | 42336478013 | 300 | 1065.1 | 875.46 | 28 | 948.25 | 814.95 | 28 |
| | | | 500 | 594.2 | 419.345 | 30 | 852.37 | 783.42 | 30 |
| | | | 700 | 348.4 | 241.267 | 30 | 518.633 | 404.516 | 30 |
| 4000 | 38 | 272903119607 | 300 | 2038.63 | 988.36 | 19 | 1889.375 | 1208.1 | 24 |
| | | | 500 | 1318 | 901.62 | 23 | 1410.65 | 1076.82 | 26 |
| | | | 700 | 1020.4 | 823.96 | 26 | 1190.83 | 1008.43 | 29 |
| 12000 | 44 | 11683458677563 | 700 | 5048.42 | 3343.37 | 20 | 5477.52 | 3712.87 | 22 |
| | | | 1000 | 4994.3 | 3597.93 | 24 | 4322.84 | 3174.13 | 26 |
| | | | 1300 | 4346.96 | 3845.87 | 26 | 4189 | 2637.23 | 29 |

## 6 Conclusion and Future Research Challenges

This work has shown the effect of chaos in improving the performance of the Grey Wolf Optimizer for continuous optimization. A new one-dimensional function has also been proposed which has shown better performance than the one being used so far. However, there still remain challenges to solving the prime factorization problem using meta-heuristics. The standard deviation in the number of iterations is of the same order as the mean iterations (in case of both functions). This shows that the performance is quite erratic within a range. This does raise concerns regarding the scalability and reliability of the methods. Secondly, a function needs to be developed that is smooth and has fewer local minima in order to increase the success ratios, specially for large semi-primes. A function was proposed in [15], which is smooth leading to better selection pressure.

$$\textbf{minimize } f(x) = |\log(N) - \log(x) - \log(y)| \quad x \in [10^{d-1}, \sqrt{N}], y \in [\sqrt{N}, 10^d - 1] \tag{37}$$

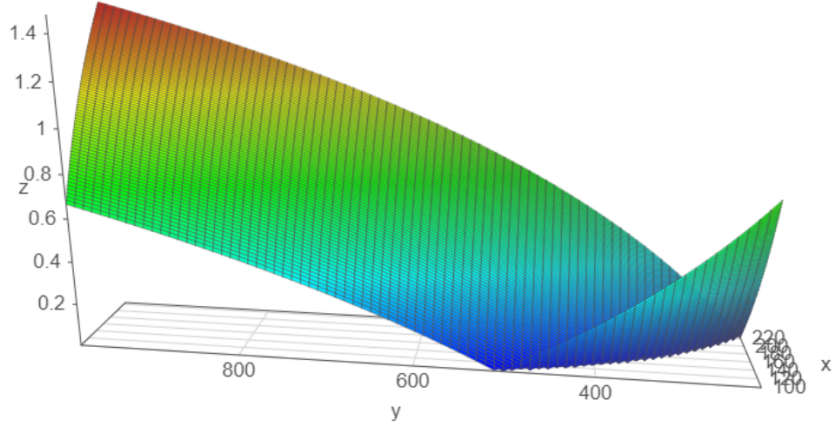

**Fig. 3** Graph of eq. (37) for $N = 50759$

However, the actual solution is located in a region having the shape of a 2-D curve. This region has a very large number of points yielding values very close to the true minimum value i.e. 0. This is evident from the graph of the function for $N = 50759$ in fig. 3. This function was also tested but yielded poor success ratios even for small numbers. Meta-heuristics are a promising method for factorizing semi-primes. Overcoming these challenges can bring about major improvements.